\documentclass[lettersize,journal]{IEEEtran}
\IEEEoverridecommandlockouts
\usepackage{cite}
\usepackage{balance}
\usepackage[caption=false,font=footnotesize]{subfig} % subfigures, IEEE-compliant
\usepackage{enumitem} % tighter lists
\usepackage{amsmath,amssymb,amsfonts}
\usepackage{graphicx}
\usepackage{subcaption} % for subfigures
\graphicspath{ {./figures/} }
\usepackage{textcomp}
\usepackage{xcolor}
\def\BibTeX{{\rm B\kern-.05em{\sc i\kern-.025em b}\kern-.08em T\kern-.1667em\lower.7ex\hbox{E}\kern-.125emX}}
\usepackage{newtxtext}
\usepackage{orcidlink}
\usepackage{changes}

\setlist{nosep,leftmargin=*} % compact itemize/enumerate, full width
\makeatletter
\g@addto@macro\normalsize{%
 \setlength\abovedisplayskip{8pt plus 2pt minus 4pt}%
 \setlength\belowdisplayskip{8pt plus 2pt minus 4pt}%
 \setlength\abovedisplayshortskip{0pt plus 2pt}%
 \setlength\belowdisplayshortskip{5pt plus 2pt minus 3pt}%
}
\makeatother

\usepackage{tikz}
\usepackage{pgfplots}
\usetikzlibrary{datavisualization}
\usetikzlibrary{positioning,arrows.meta,shapes.geometric,calc,fit}
\usetikzlibrary{fit,positioning,backgrounds}
\pgfplotsset{compat=1.17}
\pgfplotsset{
    scale only axis,
    xlabel near ticks, ylabel near ticks,
    every axis title shift=3pt
}
\pgfplotsset{
    every axis plot/.append style={
        every mark/.append style={solid}
    }
}

\definecolor{black}{RGB}{0,0,0}
\definecolor{orange}{RGB}{230,159,0}
\definecolor{light_blue}{RGB}{86,180,233}
\definecolor{green}{RGB}{0,158,115}
\definecolor{yellow}{RGB}{240,228,66}
\definecolor{blue}{RGB}{0,114,178}
\definecolor{red}{RGB}{213,94,0}
\definecolor{purple}{RGB}{204,121,167}

\usepackage{breqn} % make /left /right work with line breaks
\usepackage{mathtools} % additional math symbols
\usepackage{dsfont} % math font
\usepackage{amsfonts} % blackboard bold characters e.g. \mathbb{letters}
\usepackage{soul} % highlight text using \hl{} (requires xcolor package)
\let\oldhl\hl
\renewcommand{\hl}[2][yellow]{\sethlcolor{#1}\oldhl{#2}\sethlcolor{yellow}}

\usepackage{float}
\usepackage{stfloats}
\usepackage{multirow}
\usepackage{algorithm}
\usepackage{algpseudocode}
\usepackage{svg} % \includesvg{}
\usepackage{hyperref}

\usepackage[acronym]{glossaries}
\makeglossaries

\newacronym{int8}{INT8}{Integer 8}
\newacronym{uint8}{UINT8}{Unsigned Integer 8}
\newacronym{int16}{INT16}{Integer 16}
\newacronym{int24}{INT24}{Integer 24}
\newacronym{int32}{INT32}{Integer 32}
\newacronym{fp16}{FP16}{Floating-Point 16}
\newacronym{fp32}{FP32}{Single-Precision Floating-Point}
\newacronym{cnn}{CNN}{Convolutional Neural Network}
\newacronym{ml}{ML}{Machine Learning}
\newacronym{ai}{AI}{Artifical Intelligence}
\newacronym{dl}{DL}{Deep Learning}
\newacronym{nn}{NN}{Neural Network}
\newacronym{gpu}{GPU}{Graphics Processing Unit}
\newacronym{cpu}{CPU}{Central Processing Unit}
\newacronym{relu}{ReLU}{Rectified Linear Unit}
\newacronym{relu6}{ReLU6}{Rectified Linear Unit 6}
\newacronym{qat}{QAT}{Quantization-Aware Training}
\newacronym{ptq}{PTQ}{Post-Training Quantization}
\newacronym{hqat}{H-QAT}{Hardware-Quantization-Aware Training}
\newacronym{rqat}{R-QAT}{Rescaling- and Quantization-Aware Training}
\newacronym{ste}{STE}{Straight-Through Estimator}
\newacronym{isa}{ISA}{Instruction Set Architecture}
\newacronym{simd}{SIMD}{Single-Instruction Multiple-Data}
\newacronym{double-rounding}{VQRDMULH}{saturating Rounding Doubling Multiply returning High half}
\newacronym{mae}{MAE}{Mean Absolute Error}
\newacronym{mse}{MSE}{Mean Squared Error}
\newacronym{ema}{EMA}{Exponential Mean Average}

\newglossaryentry{flatbuffer}
{
    name=\href{https://github.com/google/flatbuffers}{FlatBuffers},
    description={is a cross platform serialization library architected for maximum memory efficiency. It allows you to directly access serialized data without parsing/unpacking it first, while still having great forwards/backwards compatibility.}
}
\newglossaryentry{integer quantized}
{
    name=Integer quantized,
    description={means weights as well as activations are quantized.}
}
\newglossaryentry{full-integer-arithmetic}
{
    name=Full-Integer Arithmetic,
    description={similar to \gls{integer quantized}. However it specifically only uses integer arithmetic during inference.}
}
\newglossaryentry{rescaling}
{
    name=Rescaling,
    description={In literature also called \textit{Requantization}. Means of transitioning quantized values from one quantization scheme to another. This might also involve an increase or decrease in bit-width}
}
\newglossaryentry{Fine-Tuning}
{
    name=Fine-Tuning,
    description={Retraining a pre-trained model to embed slight adjustments.}
}

\newglossaryentry{tflite}
{
    name={TF-Lite},
    description={TensorFlow Lite (meanwhile renamed to LiteRT)}
}

\newglossaryentry{batchnorm}{
    name={BatchNorm},
    description={A technique that normalizes input data for each layer in a neural network, by subtracting the mean and dividing by the standard deviation, to improve training stability and speed.}
}

\newglossaryentry{softmax}
{
    name={SoftMax},
    description={A function that maps a vector to a probability distribution, where each output value is between 0 and 1 and they sum to 1.}
}

\usepackage{titlesec}

\renewcommand\thesubsection{\Alph{subsection}}
\titleformat{\subsection}[runin]  % 'runin' puts the title inline
  {\bfseries}                     % formatting style (bold)
  {\thesubsection.}               % numbered label
  {0.35em}                     % spacing between label and title
  {}                             % before-code
  [:]                            % punctuation after title

\renewcommand\thesubsubsection{\arabic{subsubsection}}
\titleformat{\subsubsection}[runin]  % 'runin' puts the title inline
  {\normalfont\em}                     % formatting style (bold)
  {\thesubsubsection.}               % numbered label
  {0.3em}                          % spacing between label and title
  {}                             % before-code
  [:]                            % punctuation after title

\begin{document}

\title{ Rescaling-Aware Training for Efficient Deployment of Deep Learning Models on Full-Integer Hardware }

\author{
    Lion M{\"u}ller\orcidlink{0009-0007-5324-4294}, Alberto Garc\'{\i}a-Ortiz\orcidlink{0000-0002-6461-3864}, Ardalan Najafi\orcidlink{0000-0002-6529-4084}, Adam Fuks and Lennart Bamberg\orcidlink{0000-0003-4673-8310}
	% \ifbool{true}{
	% 	% true
	% 	\IEEEauthorblockN{
	% 		Lion M{\"u}ller\orcidlink{0009-0007-5324-4294}\IEEEauthorrefmark{2},
	% 		Alberto Garc\'{\i}a-Ortiz\orcidlink{0000-0002-6461-3864}\IEEEauthorrefmark{1},
 %            Ardalan Najafi\orcidlink{0000-0002-6529-4084}\IEEEauthorrefmark{2},
 %            Adam Fuks\IEEEauthorrefmark{2},
 %            Lennart Bamberg\orcidlink{0000-0003-4673-8310}\IEEEauthorrefmark{2}}
 %            \\
	% 	\IEEEauthorblockA{
 %            \IEEEauthorrefmark{2}\textit{NXP Semiconductors},
 %            \IEEEauthorrefmark{1}\textit{University of Bremen}}
 %        % \IEEEauthor
	% }{
	% 	% false
	% 	double blinded
	% }
    \vspace{-8mm}

    %\thanks{Lion M{\"u}ller, Ardalan Najafi, and Lennart Bamberg are with NXP Semiconductors Germany GmbH. Adam Fuks is with NXP Semiconductors USA Inc. Alberto Garc\'{\i}a-Ortiz is with University of Bremen.}

    \thanks{Lion M{\"u}ller, Ardalan Najafi, and, Adam Fuks Lennart Bamberg are with NXP Semiconductors. Alberto Garc\'{\i}a-Ortiz is with University of Bremen.}
    \thanks{This work is funded by the IPCEI ME/CT initiative by the European Union.}
    %\thanks{This work is funded by the IPCEI ME/CT initiative an is funded by the European Union Next Generation EU program, the German Federal Ministry for Economic Affairs and Energy, the Bavarian Ministry of Economic Affairs, Regional Development and Energy, the Free State of Saxony---supported by tax revenues based on the budget approved by the Saxon State Parliament---and the Free and Hanseatic City of Hamburg.}

}
\maketitle

\begin{abstract}
%% OLD longer
% Integer-only AI inference significantly enhances memory efficiency and computational throughput, making it well-suited for deployment in low-power embedded systems. Although quantization-aware training (QAT) helps mitigate accuracy degradation associated with post-training quantization, it typically overlooks the impact of integer rescaling during inference---a hardware costly operation in traditional integer-only AI inference.
% % This work shows that the cost of rescaling in integer-only inference can be significantly reduced by applying stronger quantization to the rescale multiplicands---achieving this without any degradation in model accuracy.
% This work shows that rescaling cost can be dramatically reduced post-training, by applying a stronger quantization to the rescale multiplicands at no model-quality loss. 
% To reduce the rescaling cost even further, we introduce Rescale-Aware Training, a fine tuning method that accounts for ultra-low bit-widths of the rescaling multiplicands. By emulating the integer inference path during training, the approach aligns model optimization with real deployment constraints. Experiments show that even with 8$\times$ reduced rescaler widths, accuracy is preserved through minimal incremental retraining of a small subset of weights. This enables more energy- and cost-efficient AI inference for resource-constraint edge devices.
%%% NEW SHORTER
Integer AI inference significantly reduces computational complexity in embedded systems. Quantization-aware training (QAT) helps mitigate accuracy degradation associated with post-training quantization but still overlooks the impact of integer rescaling during inference, which is a hardware costly operation in  integer-only AI inference.
% This work shows that the cost of rescaling in integer-only inference can be significantly reduced by applying stronger quantization to the rescale multiplicands---achieving this without any degradation in model accuracy.
This work shows that rescaling cost can be dramatically reduced post-training, by applying a stronger quantization to the rescale multiplicands at no model-quality loss. 
%To reduce the rescaling cost even further, 
Furthermore, we introduce Rescale-Aware Training, a fine tuning method for ultra-low bit-width rescaling multiplicands. %By emulating the integer inference path during training, the approach aligns model optimization with real deployment constraints. 
Experiments show that even with 8$\times$ reduced rescaler widths, the full accuracy is preserved through minimal incremental retraining. This enables more energy-efficient and cost-efficient AI inference for resource-constrained embedded systems.
\end{abstract}

\vspace{-3mm}

\section{Introduction}

\begin{figure}[!t]
    \centering
    %\includesvg[width=0.9\linewidth]{figures/SVG/rescaling_unit-3.svg}
    \includegraphics[width=0.9\linewidth]{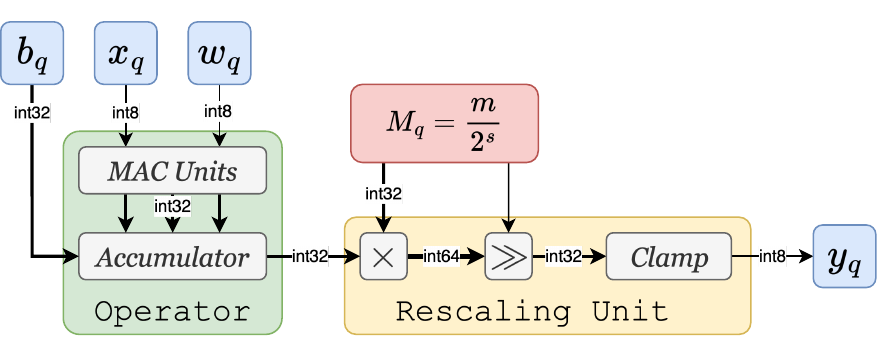}
    %\caption{Schematic visualization of the rescaling unit (yellow) within the context of an integer-only operation. Next to the individual parts involved in the process the bit-width of the data paths are shown.}
    %\caption{Schematic visualization of full-integer arithmetic including bit-widths as proposed in~\cite{jacob2017}.}
    \caption{Visualization of full-integer arithmetic  proposed in~\cite{jacob2017}.}
    %\vspace{-4mm}
    \label{fig:rescaling-unit}
\end{figure}

Hardware-efficient full-integer-quantized AI inference is the de-facto standard in today's embedded systems.  %, driven by their benefits in computational efficiency and reduced memory usage. %However, these models often face challenges in maintaining accuracy, especially under aggressive quantization strategies~\cite{jacob2017}. 
A crucial step of full-integer quantization is the rescaling of high bit-width accumulators to low bitwidth outputs.
%Rescaling involves multiplying the 32-bit integer accumulator by a fractional number $M \in (0,1]$, followed by casting the result to a lower-bitwidth integer. T
This operation is implemented in integer-only hardware  by multipying the 32-bit accumulator with a wide unsigned integer, $m$, followed by a shift-right-and-round and casting to 8-bit~\cite{jacob2017}. 
%using integer arithmetic by expressing $M$ as a dyadic approximation $M_q = m \cdot 2^{-s}$, where $m$ and $s$ are unsigned integers of 32-bit and 8-bit width, respectively.
Consequently, commercial embedded integer neural processing units (NPUs) such as \cite{neutron_ref} integrate multiple wide multipliers for data-parallel processing (see Fig.~\ref{fig:rescaling-unit})--- causing significant hardware cost. %and thereby reducing overall efficiency.
%In particular, the wide multiplication
%\footnote{\href{https://developer.arm.com/documentation/102684/0000/Functional-description-of-the-Arm--Ethos-U85--NPU}{Ethos-U85 Documentation}}

One approach to reduce rescaling overhead is to restrict the rescaler to a power-of-two, allowing the operation to be implemented using simple bit-shifting~\cite{li2020, elhoushi_deepshift_2021, przewlockarus2022, li_denseshift_2023}. However, this requires coarse discretization of the quantization scales, which can lead to noticeable degradation in model accuracy.

We aim to optimize the cost-quality trade-off by reducing the bit-width of $m$ from 32 bits to a smaller value. This also decreases the width of the multiplication output,  reducing the cost of the subsequent shift-and-cast operation. %and potentially allowing for narrower representation of the shift parameter $s$. 
We show that one can minimize the rescaling width to 8 bit, from today's 32 bit, without impairing model accuracy.

%To further reduce the cost---by means of an extremely low-bitwidth ($<8\,b$) quantized $m$---without model quality degradation, we propose a fine-tuning method called \textit{Rescale-Aware Training}. This method emulates the quantization errors of the rescale hardware during the forward pass, while ensuring proper gradient propagation during the backward pass using Straight-Through Estimation for rounding operations. By leveraging the inherent noise resilience of deep neural networks, our approach enables an $8\times$ reduction in the bit-width of the rescaler multiplicand while preserving model accuracy, all within a reasonable retraining effort.
To enable even lower bitwidths for $m$, we propose a fine-tuning method called \textit{Rescale-Aware Training}. This method emulates the quantization errors of the rescale hardware during the forward pass. Proper gradient propagation during the backward pass is ensured by Straight-Through Estimation. %for rounding operations. 
By leveraging the inherent noise resilience of deep neural networks, our approach enables an $8\times$ reduction in the bit-width of the rescaler multiplicand while preserving model accuracy, all within a reasonable retraining effort.
% By harnessing the adaptability (noise resilience) of deep neural networks, we can drastically reduce the rescaler multiplicand bit-width by $8\times$ while maintaining model accuracy within a reasonable retraining time.

%\vspace{-4mm}
\subsection{Related Work}

Integer-only arithmetic for efficient AI inference on resource-constrained devices was introduced by Jacob et al.~\cite{jacob2017}, along with \gls{qat}, %to simulate quantization effects during training. Compared to \gls{ptq}, \gls{qat}
enabling full-integer quantization with 8-bit weights and 32-bit rescaling while maintaining accuracy.

Subsequent works refined quantization techniques to improve accuracy and hardware efficiency. Nagel et al.~\cite{nagel2019} proposed weight equalization and bias correction to reduce quantization-induced bias. Li et al.~\cite{li2020} introduced Additive Powers-of-Two (APoT) weight quantization, enabling shift-and-add based, low-cost  dot-product. %and improved stability via weight normalization
Efficient rescaling has been addressed by training all quantization thresholds under Power-of-Two (PoT) constraints~\cite{jain2020}\@. This was extended with hardware-compliant schemes with better resolution near zero to improve accuracy~\cite{przewlockarus2022}.

%Other directions include activation function modifications, such as Bounded ReLU (BReLU) by Zhao et al.~\cite{zhao2020}, and accumulator-aware quantization (A2Q) by Colbert et al.~\cite{colbert2023}, which constrains accumulator bit-width to prevent overflow.

Existing power-of-two methods have advanced the field of integer-only inference, by a fixed trade-off of cost for model quality. A fine-grain cost-quality trade-off in rescaling remains not explored in sufficient detail. This work addresses that gap by proposing \emph{Rescale-Aware Training}, a technique that adapts pre-trained models to low-bitwidth rescaling multiplicands without compromising accuracy.

\vspace{-2mm}
\section{Methodology}
\vspace{-1mm}
Our implementation is based on the LiteRT framework, which compiles models into a hardware-agnostic intermediate representation. We modify this framework to enable accurate inference modeling under specific rescaler configurations. The same approach is used to implement rescale-aware training.

\vspace{-3mm}
\subsection{Preliminaries}
This section formalizes the quantized inference setting and introduces the relevant assumptions, quantization schemes, and computational models.

\subsubsection{Quantization Scheme}~We assume an integer-only inference pipeline with the following quantization configurations:
\begin{itemize}
  \item \textbf{Weights:} Symmetric, per-channel, 8-bit, scale $S_w$
  \item \textbf{Activations:} Affine, 8-bit, scale $S_x$, zero-point $Z_x$
  \item \textbf{Bias:} Symmetric, per-channel, 32-bit, scale $S_x \cdot S_w$
\end{itemize}
This format aligns with all widely adopted deployment practices (e.g., LiteRT, TOSA, Executorch) and enables efficient integer-arithmetic hardware.

\subsubsection{Integer Quantized Inference}~Let $\vec x_q$, $y_q$, and $\vec w_q$ denote the 8-bit quantized input/output activations and weights, with scales $(S_x, S_y, S_w)$ and zero-points $(Z_x, Z_y)$. The core scaled-and-biased dot-product operation (used to realize matrix multiplications, convolutions, etc. in AI) is computed as:

\begin{equation}
y_{q} = \mathrm{Sat}_{\mathrm{int8}}( \lfloor M_q \underbrace{\sum_i (x_{q,i} \cdot w_{q,i}) + b_{\mathrm{eff}}}_{\textit{a}_q} \rceil )
\end{equation}

Here, $b_{\mathrm{eff}}$ is the effective bias, incorporating both the neural bias and the impact of input/output zero-points as described in~\cite{jacob2017}. The 32-bit symmetrically quantized accumulator, $a_q$, holds the biased results and the dot-product at scale $S_x \cdot S_{w}$.
To cast the accumulator output to the desired scale $S_y$, a rescaling factor $M = \frac{S_x \cdot S_{w}}{S_y}$ is applied. Since the bit-width is reduced from 32 to 8 bits, $M$ falls within the range $(0, 1]$.

% \begin{equation}
% M = \frac{S_x \cdot S_{w}}{S_y}\text{.}
% \end{equation}
The final saturation to the int8 range not only casts the result to the target format but also implements activation functions that are a mere clamping (e.g., ReLU, ReLU6), commonly used in integer-only inference.

\subsubsection{Quantization of the rescale factor}
\label{sec:quantization_of_rescale_factor}
~The dyadic quantization of the rescale factor into an integer multiplier and associated right-shift-and-round is derived from \cite{jacob2017}:
\begin{equation}
    \label{eq:normalized_rescale_factor}
    M_q =  m \cdot 2^{-s}
\end{equation}

In \cite{jacob2017}, $m$ is a 32-bit integer and $s$ is chosen to have a leading 1-bit in $m$ to maximize the effective precision gained. 
% Thus:
% % \begin{align}
% %     %\label{eq:normalized_rescale_factor}
% %     s &=  \lceil 31-\log_2(M) \rceil \nonumber \text{,} \\
% %     m &= \lfloor M\cdot 2^s \rceil \text{.} \label{eq:old_m}
% % \end{align}
% \begin{equation}
%     %\label{eq:normalized_rescale_factor}
%     s =  \lceil 31-\log_2(M) \rceil \text{,} \ \  m = \lfloor M\cdot 2^s \rceil \text{.} \label{eq:old_m}
% \end{equation}  
We propose keeping the multiplier width as $k<32b$, investigating how small we can go without loss of accuracy.

% Turning to actual implementations of rescale factor quantization, like in LiteRT, we see a different way of calculating the multiplier and shift. Deducing them from the fractional part and the exponent of the IEEE 754 floating point representation \eqref{eq:ieee-754}
This work proposes a computationally efficient way to obtain the $k$-bit value $m$ and the shift amount $s$ compared to~\cite{jacob2017}, to speed up our proposed rescale-aware training. %outlined in Subsection \textit{C}.
The real-valued IEEE 754 floating-point  rescale value $M$ is represented by a fractional part (mantissa) and exponent as:
%\begin{equation}
    %\label{eq:ieee-754}
    $r \simeq (1 + fr) \cdot 2^{\textit{exp}}$.
%\end{equation}
The multiplier $m$ can be obtained without arithmetic operations by appending the $k$ most significant bits of mantissa $fr$ to a leading 1 (hidden bit).
Similarly, $s$ is obtained from \textit{exp} by adding a simple integer constant.
%\begin{equation}
%    \label{eq:dyadic-rescaler}
%    M_q = \underbrace{ \left\lceil(1+f) \cdot 2^k\right\rfloor }_{m} \cdot 2^{e-k}
%\end{equation} 

Note that even for small $k$, $s$ will be positive in all cases, as $m\geq 1$ and $M\leq 1$. Thus, a narrow \textit{uint} is used for $s$. With low-$k$ inference, $s$ can use as few as $\log_2(32+k-8)$ bits. The multiplication results of the accumulator and $m$ is of width $32+k$, shifted right by at most $32+k-8$ to fully exploit the 8-bit output range.
Consequently, reducing $k$ not only lowers the complexity of the multiplication stage but also simplifies the subsequent shift and rounding operations.

\subsubsection{Rescale Error Model}
~The quantization error of the quantized accumulator value before rescaling, $a_q$, relative to the true value $a$, is obtained by dequantization the value to $\tilde a$ (i.e., multiplying with the quantization scale) as
\begin{equation}
\label{eq:acc_quant_error}
    \epsilon_a = a- \tilde a = a - S_x S_w a_q\text{.}
\end{equation}
This error comes from the weight, input, and bias quantization.
During rescaling, the quantized accumulation result is multiplied by the quantized rescaler $M_q$, defined in Eq.~\eqref{eq:normalized_rescale_factor}. %The operation itself is executed in two steps whereby the right-shift operation is also rounding the result to the nearest \textit{int8}. 
After rescaling the quantized values use scale $S_y$.
Dequantizing, to $\tilde y$ allows obtaining the output error:
\begin{equation}
    \label{eq:error_after_rescaling}
    \epsilon_y = y -\tilde  y = y - S_y\cdot \mathrm{Sat}_{\mathrm{int8}}\left(\lceil M_q a_q \rfloor \right)
\end{equation}

Comparing the dequantized outputs before and after rescaling, yields the error introduced by the rescaling operation: 
%%%% STOPPED %%%
% \begin{align}
%     \label{eq:res_error}
%     \epsilon_r & = \epsilon_a - \epsilon_y = \tilde y - \tilde a \text{,} \\
%     \label{eq:res_error_final}
%     & = S_y a_q (M_q - M) + S_y \delta_r \text{.}
% \end{align}
\begin{equation}
    \label{eq:res_error_final}
    \epsilon_r = \epsilon_a - \epsilon_y = \tilde y - \tilde a = S_y a_q (M_q - M) + S_y \delta_r 
\end{equation}
%     \label{eq:res_error}
%     \epsilon_r & = \epsilon_a - \epsilon_y = \tilde y - \tilde a \text{,} \\
%     \label{eq:res_error_final}
%     & = S_y a_q (M_q - M) + S_y \delta_r \text{.}
% \end{align}
% To evaluate the worst-case error introduced by the rescaling operation, we assume that saturation does not affect the result (i.e., $a=y$). The rationale is that active saturation within valid quantization ranges typically clips the error to zero---particularly in cases such as ReLU activation, where zero is represented exactly due to the presence of a zero-point.
Active saturation under valid quantization ranges ties the error typically to 0 (ReLU activation as 0 is always quantized error-free).
Thus, we assume no saturation (i.e., $a=y$), to consider the worst-case error.
The derived formulation of the rescale error $\epsilon_r$  consists of two parts.
One due to the discrepancy between the ideal rescaling factor $M$ and its dyadic approximation $M_q$. %applied to the quantized accumulator. 
After rescaling, this error is scaled by the output quantization scale $S_y$ times the accumulator value $a_q$, and is bounded by ${|M_q-M|\cdot \max(|\tilde y|)}$.
The second component originates from the rounding of the right-shift operation (\textit{cf.} Algorithm~\autoref{alg:multiply-by-quantized-multiplier}). This rounding error is uniformly distributed in the range $(-\frac{S_y}{2}, \frac{S_y}{2})$.

For reduced rescale bit-widths, the component of the error associated with quantization of the rescale factor exhibits exponential growth, with its magnitude distributed proportionally to the accumulator. This occurs when ${|M_q-M|\cdot\max|S_ya_q| > \frac{S_y}{2}}$, caused by more severe quantization of the rescale factor. In contrast, when $|M_q - M|\rightarrow0$, the total rescale error is instead dominated by the rounding/quantization error from going down to 8-bit. 
%%%%%%%%%%%%%%%%%%%%%%%%%%%%%%%%%%%%
% END of Preliminaries
%%%%%%%%%%%%%%%%%%%%%%%%%%%%%%%%%%%%
\vspace{-3mm}
\subsection{Validation (Emulate Rescaling Step)}
To validate the proposed method, we guarantee output parity between the forward path during retraining and quantized inference. This is achieved by recompiling LiteRT with modified reference kernels. %LiteRT is emulating a special instruction (\verb|VQRDMULH|)
%(\verb|VQRDMULH|\footnote{\href{https://developer.arm.com/documentation/dui0473/m/neon-instructions/vqrdmulh--by-vector-or-by-scalar-}{https://developer.arm.com/documentation/dui0473/m/neon-instructions/vqrdmulh--by-vector-or-by-scalar-}}) 
%by default. This was disabled as the instruction is specifically tailored to operate on similar saturated numbers of the same datatype (e.g. \verb|int32|). 
In detail, the global routine to quantize floating-point rescale factors is altered to reflect a specific rescaling bit-width. Similarly, the routine to perform the rescaling operation is adapted (\textit{cf.} Algorithm~\autoref{alg:multiply-by-quantized-multiplier}).

\begin{algorithm}[t]
\footnotesize
\caption{MultiplyByQuantizedMultiplier}
\label{alg:multiply-by-quantized-multiplier}
\begin{algorithmic}[1]
    \Ensure Gradient propagation of all rounding operations
    \Function{MultiplyByQuantizedMultiplier}{x, multiplier, shift}
        % \Comment{Round() and Floor() propagate gradients by STE}
        \State $rnd \gets 2^{(\text{shift}-1)}$ \Comment enable round-to-nearest
        \State $mul \gets x \times \text{multiplier} + rnd$
        \State $res \gets \text{Floor}(mul \times 2^{-\text{shift}})$
        % \State  \gets \text{Floor}(shr)$
        \State \Return $res$
    \EndFunction
\end{algorithmic}
\end{algorithm}
\vspace{-2mm}

\subsection{Rescale-Aware Finetuning}
\label{sec:fine-tuning}
To support effective training under quantization constraints, our method adopts a rescale-aware training approach that emulates integer inference using floating-point arithmetic. This ensures consistency with the behavior of LiteRT's quantized deployment path while maintaining accurate differentiability. The key considerations and mechanisms of this training procedure are outlined below.

\subsubsection{Hardware Modeling}
\label{sec:hardware-assumptions}
Although LiteRT compiles models into a generic intermediate representation, it implicitly assumes specific hardware behaviors during quantized inference. To accurately replicate these behaviors in the training loop, we adopt the following assumptions:

\begin{itemize}
    \item \textbf{Accumulator Precision:} 32-bit accumulators are used. %throughout kernel operations.  %, which themselves remain unaltered.
    \item \textbf{Rounding Strategy:} Rounding follows a hardware-efficient round-to-nearest (round-half-up) convention.
    \item \textbf{Output Saturation:} Rescaling is performed with int64 intermediate results to avoid overflow. The result is stored in int32 and then saturated to the output data type (e.g., int8).
\end{itemize}

\begin{figure}[t]
    \centering
    %\includesvg[width=0.85\linewidth]{figures/SVG/finetuning-graph-conversion_new.svg}
    \includegraphics[width=0.88\linewidth]{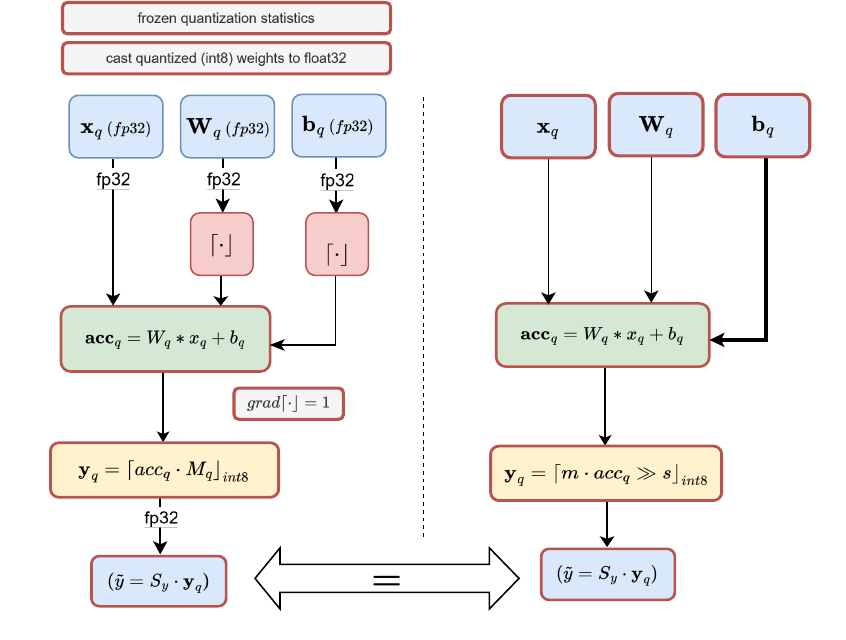}
    \caption{Left: training graph with emulated integer-only arithmetic. Right: integer quantized inference graph.}

    \label{fig:finetuning-graph-conversion}
\end{figure}

\subsubsection{Emulation of Integer Inference during Training}~
Training under these hardware assumptions requires floating-point emulation of integer arithmetic (shown in \autoref{fig:finetuning-graph-conversion}). We derive a quantization-aware training loop that mirrors integer inference behavior while supporting gradient-based updates:

\begin{itemize}
    \item \textbf{Weight Preparation:} 
    % Pre-trained LiteRT models are used as initialization. Integer weights are cast to floating-point format without dequantization. Since quantization scales are assumed constant, the corresponding rescale factors are also treated as fixed.
    Pre-trained LiteRT models are used for initialization. Integer weights are cast directly to floating-point format without undergoing dequantization. Given that quantization scales are assumed to be fixed, the corresponding rescale factors are likewise treated as constant.
    \item \textbf{Numerical Stability:} Accumulations and rescaling operations are performed in float64 to match the precision of int32. This prevents rounding instabilities that may arise due to the limited 24-bit mantissa of float32.
    \item \textbf{Fake Quantization:} All parameters are subjected to fake quantization during the forward pass. %to closely emulate the behavior of the final quantized model. 
    Thereby, the outputs produced during training are representative of those observed during inference.
    \item \textbf{Gradient Flow via STE:} Since rounding is a non-differentiable, discretized operation, we use the Straight-Through Estimator (STE) \cite{bengio2013} during backpropagation. %This allows effective backpropagation despite the presence of discrete operations.
\end{itemize}
%\vspace{-3mm}
\subsection{Model Re-Deployment}
After fine-tuning, the updated floating-point weights are re-quantized by rounding them to integers. These updated values are directly substituted into the LiteRT flatbuffer model, effectively replacing the original quantized weights with those optimized through training, while keeping the original quantization parameters.
\vspace{-2mm}
\section{Experimental Results}

\subsection{Post-Training Rescaler Bit-Width Reduction}
First, we quantify model robustness to post-training quantization of the rescale-multiplicand bit-widths. 
%Specifically, we investigate the threshold bit-width at which model performance begins to degrade.
This point not only serves as a reference point for our fine-tuning method, but also defines for hardware architects the cost reduction opportunity over the state-of-the-art even without model fine-tuning. 

We consider standard efficient-AI models trained on the ImageNet1000 classification task. We define the degradation point as the rescaling bit-width at which the classification accuracy declines by more than 0.5\%. The results are presented in \autoref{tab:lower-bound-analysis}.

% We consider the standard Edge-AI ImageNet1000-classification models as benchmarks, and define the degradation point as the rescaling bit-width at which the classification accuracy degrades by over 2\%. The results are shown in \autoref{tab:lower-bound-analysis}.

%Shown exemplary for the \textit{EfficientNet-Lite0} model in \autoref{fig:performance-recovery} (lower-bound analysis), the model is barely affected by the increased rescaling noise, although the rescaling bit width has been greatly reduced to 8 bits. Only under extreme reduction is an effect visible that quickly leads to a breakdown.
%Similar behavior is observed for different model architectures of comparable parameter size, shown in \autoref{tab:lower-bound-analysis}
\begin{table}[t]
    \centering
    \caption{Impact of post-training rescaler bit‑width reduction on different LiteRT ImageNet classification models.}
    \resizebox{\columnwidth}{!}{%
    \begin{tabular}{l c c c c c c}
        \hline
        %--- first header row -------------------------------------------------
        \multirow{2}{*}{\textbf{Model}} &
        \textbf{Base} &
        \multicolumn{5}{c}{\textbf{Accuracy per rescaler bit‑width}} \\
        \cline{3-7}
        %--- second header row ------------------------------------------------
        & \textbf{Accuracy} & 8 & 6 & 5 & 4 & 3 \\
        \hline \hline
        %--- data rows --------------------------------------------------------
        % EfficientNet-Lite0 &
        % 68.32\% & 68.45\% & 67.17\% & 66.80\% & \textbf{62.92\%} & 20.48\% \\
        EfficientNet-Lite0 &
        71.28\% & 70.93\% & \textbf{69.83\%} & {69.20\%} & 65.39\% & 21.72\% \\
        % MobileNet-V1 &
        % 68.01\% & 68.11\% & 67.68\% & \textbf{64.54\%} & 60.05\% & 4.84\% \\
        MobileNet-V1 &
        70.51\% & 70.51\% & 70.41\% & \textbf{67.24\%} & 63.67\% & 05.11\% \\
        % MobileNet-V2 &
        % 69.07\% & 69.08\% & 68.55\% & \textbf{66.98\%} & 51.12\% & 23.76\% \\
        MobileNet-V2 &
        71.09\% & 71.10\% & 70.82\% & \textbf{69.28\%} & 54.55\% & 23.26\% \\
        \hline
    \end{tabular}
    }
    %vspace{-4mm}
    \label{tab:lower-bound-analysis}
\end{table}

Our findings indicate that reducing the rescaling width to 8 bits (representing a 4$\times$ reduction over the state-of-the-art) does not compromise model accuracy in all cases. %Notably, performance degradation only emerges when the rescaling width is reduced below 6 bits.

\subsection{Rescale-Aware Fine Tuning}
This subsection analyzes the effect of rescale-aware fine-tuning of model weights, using the \textit{EfficientNet-Lite0} architecture as a case study. We investigate both, the proportion of weights that were modified during the process and the mean absolute deviation of the updated weight values relative to those in the original quantized model.

All experiments were conducted using Stochastic Gradient Descent (SGD) without any modification to the default learning rate provided by TensorFlow. Although the weights were directly taken from the quantized model—thus retaining large absolute values due to the absence of dequantization—the optimizer proved stable and effective.

The assumption that only minimal weight changes would be required, is confirmed by the empirical results. Weight changes following 2 epochs of retraining for a 4-bit rescaler are minor. Specifically, Table~\ref{tab:eval_4_bit} indicates that merely 0.66\% of the weights were updated, with the average absolute change remaining below 0.71\%. Despite the limited extent of these changes, they were sufficient to yield a substantial improvement in classification accuracy---approximately 6\% after two epochs, resulting in full accuracy recovery.

\begin{table}[t]
    \centering
    \caption{EfficientNet-Lite0 with a 4-bit rescaler. 
    }
    % \begin{tabular}{l l c c  c}
    %     \hline
    %     %--- first header row -------------------------------------------------
    %     \multicolumn{2}{l}{\multirow{2}{*}{\textbf{Evaluation}}} &
    %     \multirow{2}{*}{\textbf{Accuracy}} & 
    %     \multicolumn{2}{c}{\textbf{Weight Change [\%]}} \\ 
    %     \cline{4-5}
    %     %--- second header row (sub‑headings) ---------------------------------
    %     & & & ratio & Abs. diff. \\
    %     \hline \hline
    %     %--- data rows --------------------------------------------------------
    %     LiteRT & (un-altered) & 71.28\% & & \\ 
    %     \hline
    %     Emulated & baseline & 65.39\% &  & \\ 
    %     LiteRT   & baseline & 65.42\% & & \\  
    %     \hline
    %     Emulated & re-trained (2 epochs) & 71.03\% &  \multirow{2}{*}{0.66} & \multirow{2}{*}{0.71} \\
    %     % Emulated & re-trained (2 epochs) & 71.03\% &  \multirow{2}{*}{0.0085} & \multirow{2}{*}{0.0017} \\
    %     LiteRT   & re-trained (2 epochs) & 71.62\% &  & \\ 
    %     \hline
    % \end{tabular}
    \begin{tabular}{l c c c}
    \hline
    %--- first header row -------------------------------------------------
    \multirow{2}{*}{\textbf{Evaluation}} &
    \multirow{2}{*}{\textbf{Accuracy}} &
    \multicolumn{2}{c}{\textbf{Weight Change [\%]}} \\ 
    \cline{3-4}
    %--- second header row (sub‑headings) ---------------------------------
    & & ratio & Abs. diff. \\
    \hline \hline
    Original & 71.28\% & & \\ 
    \hline
    Post-training & 65.39\% & & \\ 
    \hline
    re-trained (2 epochs) & 71.62\% & 0.66 & 0.71 \\ 
    \hline
    \end{tabular}
    %\vspace{-3mm}

    \label{tab:eval_4_bit}
\end{table}

%As illustrated in Figure~\ref{fig:rel-weight-value-change-2epoch},  % THERE WERE TWO MORE CITATIONS REMOVED FOR SPACE BELOW: zhou2017, choi2018
All weight updates observed after a single epoch of training were of magnitude ±1. 
Furthermore, only 50\% (25/50) of the layers containing weights were affected---predominantly those in the later stages of the network. This observation aligns with prior findings in aggressive quantization approaches~\cite{hubara2016}, which frequently exclude the first layer from quantization due to its heightened sensitivity.
% Although changes were sparse, they affected sparsity by introducing non-zero values where there were previously zeros (see Figure~\ref{fig:rel-weight-value-change-2epoch}). After one epoch of training, all weight updates were ±1 in magnitude.
% Furthermore, only about 50\% of the layers with weights (25 out of 50) were affected—primarily the later layers. This pattern is consistent with prior findings in aggressive quantization strategies~\cite{hubara2016, zhou2017, choi2018}, which often exclude the first layer from quantization due to its sensitivity.
% \begin{figure}[t!]
%     \raggedleft

%     %\begin{subfloat}
%     %     \centering
%     %     \includesvg[width=1\linewidth]{figures/SVG/weight_diffs_absolute_efficientNetLite0_epoch2_3_0p.svg}
%     %     \caption*{\footnotesize Absolute value difference $>3\%$ after 2 epochs}
%     %     \label{fig:abs-weight-change-2epoch}
%     %\end{subfloat}
%     %\vspace{1mm}
%     \begin{subfloat}
%         \centering
%         \includesvg[width=1\linewidth]{figures/SVG/weight_diffs_relative_efficientNetLite0_epoch2_3_0p.svg}
%         \caption*{\footnotesize Relative value difference $>3\%$ after 2 epochs fine-tuning}
%         \label{fig:rel-weight-value-change-2epoch}
%     \end{subfloat}
%     \caption{EfficientNet-Lite0 weight changes for 4-bit rescaler. %Layers with at least 3\% of weight change are shown.
%     }
%     \label{fig:main}
% \end{figure}
A second epoch revealed additional adaptivity in the upper layers and introduced minor modifications in three more layers. In particular, the first depth-wise convolutional layer exhibited consistent weight changes of ±2 across a substantial number of parameters. %These updates led to an overall increase in sparsity by approximately 1.2\%. 
Overall, we notice for this layer a reduced number of zero-valued weights (sparsity) as well as a shift in mean value.
%In addition, the resulting shift in the mean is clearly visible in Figure~\ref{fig:abs-weight-change-2epoch}, marked by the red dashed line. 
This suggests a compensatory mechanism that addresses the mean activation shift introduced during rescaling. Such dynamics are consistent with the hypothesis of biased output distributions in small-kernel layers due to quantized weights, as discussed by Finkelstein et al.~\cite{finkelstein2019}.

% A second epoch revealed further adaptivity in these upper layers and introduced minor changes in three additional layers. Notably, the first depth-wise convolutional layer exhibited a consistent ±2 change across many weights (notably, these weight changes also led to a slight increase in sparsity—--approximately 1.2\%.), producing a clear shift in the mean, marked by the red dashed line in Figure~\ref{fig:abs-weight-change-2epoch}. This suggests a compensation mechanism for the mean activation shift introduced during rescaling.
% %, as discussed in Section~\ref{sec:Error_Propagation}. 
% Such behavior aligns with the hypothesis of biased output distributions in small kernel layers due to quantized weights, as discussed by Finkelstein et al.~\cite{finkelstein2019}.

An alternative explanation is that the network compensates for systematic output bias through weight adjustments, similar in spirit to the Incremental Bias Correction method, which shows strong results with few calibration samples (typically 8–64). While bias correction explicitly adjusts layer biases, the observed behavior here suggests that incremental weight fine-tuning may achieve similar ends indirectly.

Further insight is obtained by analyzing the relative changes in weight values. After one epoch of training, the majority of weight updates exhibit a magnitude of ±1, suggesting that most updates originated from previously zero-valued (i.e., ineffective) weights.
\vspace{-1mm}

% \begin{table}[!ht]
%     \centering
%     \begin{tabular}{cccc}
%         \toprule
%         Epoch & \% Weights Updated & \% Value Change & Accuracy (\%) \\
%         \midrule
%         0 &  &  & 62.92 \\
%         1 & 0.0041 & 0.0009 & 67.18 \\
%         2 & 0.0085 & 0.0017 & 68.86 \\
%         \bottomrule
%     \end{tabular}
%     \caption{Per-epoch weight changes during incremental training of \textit{EfficientNet-Lite0} on 4-bit rescaler quantization.}
%     \label{tab:results-weight-change}
% \end{table}

\vspace{-1mm}
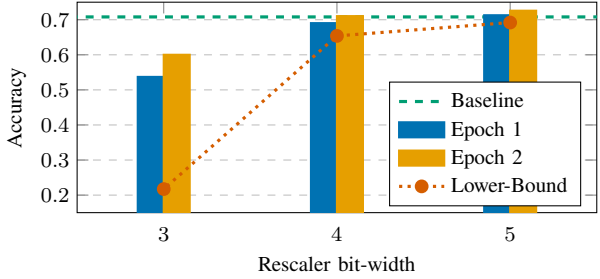
\begin{figure}[t]
    \centering
    \begin{tikzpicture}
\footnotesize
    \begin{axis}[
        xlabel={Rescaler bit-width},
        ylabel={Accuracy},
        width=0.78\linewidth,
        height=2.8cm,
        xmin=2.5, xmax=5.5,
        xtick={3,4,5},
        ymin=0.15, ymax=0.75,
        ytick={0.2,0.3,...,0.7},
        legend pos=south east,
        legend style={cells={anchor=west}, font=\footnotesize},
        ymajorgrids=true,
        grid style=dashed,
        xticklabel style={yshift=-0.1cm},
        bar width=0.15
    ]
    
    % Baseline reference line
    \addplot[dashed, very thick, green]
        coordinates {(2, 0.7080) (6, 0.7080)};
    \addlegendentry{Baseline}
    
    % Keras epoch 1
    \addplot[ybar, draw=blue, fill=blue, thick, bar shift=-0.075, area legend]
        table[x=bitwidth, y=accuracy, col sep=space]
        {data/eval-keras-retrained-1epoch.data};
    \addlegendentry{Epoch 1}
    
    % Keras epoch 2
    \addplot[ybar, draw=orange, fill=orange, thick, bar shift=0.075, area legend]
        table[x=bitwidth, y=accuracy, col sep=space]
        {data/eval-keras-retrained-2epochs.data};
    \addlegendentry{Epoch 2}
    
    \addplot[dotted, very thick, red, mark=*, restrict x to domain=3:5]
        table[x=bitwidth, y=accuracy, col sep=space]
        {data/eval-tflite.data};
    \addlegendentry{Lower-Bound}
    
    \end{axis}
\end{tikzpicture}
    \vspace{-2mm}
    \caption{EfficientNet-Lite0 accuracy recovery.} %across rescaler bit-widths, with baseline and lower-bound references.}
    \vspace{-2mm}
    \label{fig:performance-recovery}
\end{figure}
\vspace{-1mm}
\subsection{Evaluation of Performance Recovery}
\label{sec:results-performance-recovery}

% Given the \textit{almost} output parity of the Keras inference model and reference (LiteRT testbench) model and the lower bound of rescaler bit-width, the retraining is evaluated for performance recovery. For that the \textit{EfficientNet-Lite0} model was succesfully retrained on rescaler bit-widths of 4-bit and 5-bit (see \autoref{fig:eval_tflite_retrained_1epoch}). Evaluations on the testbench showed that the performance after one epoch almost reached back up to the original baseline accuracy in the 4-bit case. After a second epoch the performance gap was not only closed but slightly exceeded compared to the baseline of $68.32\%$. Although, this out-performance of the baseline might be an overfitting phenomena in regards to the potential differences in the training or even validation datasets used here and by the developers of the original model.

% The training on a 3-bit rescaler, which had a drastic performance degradation (compare \autoref{fig:eval-tflite}) has improved significantly during a single epoch of training. However, an anomaly between the evaluation results of Keras and LiteRT was observed, which can not be explained and might be accounted to the drastic reduction and recovery of model performance.

Building on the near-output parity between the integer-only arithmetic emulated infernce path and the customized LiteRT implementation, as well as the previously established lower bounds on rescaler bit-width, this section evaluates the effectiveness of retraining in restoring model performance.

% \begin{table*}[!b]
%     \centering
%     \begin{tabular}{l*{12}{c}}
%         \hline
%         %--- first header line ------------------------------------------------
%         & \multicolumn{12}{c}{(Rescaling) Multiplicand Bit‑Width} \\
%         \cline{2-13}
%         %--- second header line (the actual bit‑width values) ----------------
%         & 32 & 24 & 20 & 16 & 12 & 8 & 7 & 6 & 5 & 4 & 3 & 2 \\ 
%         \hline\hline
%         %--- data rows --------------------------------------------------------
%         Area [$\mu\text{m}^2$] &
%         870.31 & 660.56 & 572.77 & 455.96 & 339.33 & 222.61 & 217.87 & 164.92 & 157.50 & 107.50 & 97.37 & 34.42 \\
%         \hline
%         Power [$\mu\text{W}$] &
%         848 & 616 & 578 & 439 & 303 & 171 & 157 & 114 & 96 & 60 & 45 & 10 \\
%         \hline
%         Reduction [\%] &
%         0 & 24.1 & 34.2 & 47.6 & 61.0 & 74.4 & 75.0 & 81.1 & 81.9 & 87.7 & 88.8 & 96.1 \\
%         \hline
%     \end{tabular}
%     \caption{Results of chip‑area/power (savings) from synthesis of a signed multiplier with different second (rescaling) multiplicand bit‑widths in a commercial 16nm technology. The other multiplicand is fixed at 32 bits.}
%     \label{tab:synthesis}
% \end{table*}

The \textit{EfficientNet-Lite0} model was successfully retrained using rescaler bit-widths of 4 and 5 bits. As shown in Figure~\ref{fig:performance-recovery}, performance significantly improved even after a single epoch of rescale-aware training. In the 4-bit case, the accuracy nearly reached the original baseline after the first epoch. Following a second epoch, the accuracy not only closed the gap but slightly exceeded the original baseline accuracy of 71.28\%.

This marginal improvement over the baseline may be attributed to further training or subtle differences in the training and validation datasets used in this study versus those used during the original model's development. Therefore, any interpretation of out-performance should be made with caution.
In the more aggressive 3-bit case—where performance degraded substantially after quantization (see Figure~\ref{fig:performance-recovery})—retraining improves accuracy but can no longer recover the original quality.
% , showing that standard power-of-two quantization schemes cannot suffice

% However, an unexpected discrepancy was observed between evaluation results from the emulated integer inference and the LiteRT inference path. The cause of this anomaly remains unclear and may be related to the severity of the quantization or recovery process.

% Note: After retraining, the resulting model is specialized to the specific rescaler bit-width and associated hardware assumptions used during fine-tuning. These constraints are discussed in detail in Section~\ref{sec:hardware-assumptions}.

Note: After retraining, the model will only reach its maximum accuracy for the specific rescaler bit-width or hardware implementation employed during fine tuning. The assumptions made for these experiments were presented in \ref{sec:hardware-assumptions}.

\vspace{-1mm}
\subsection{Hardware efficiency}

To assess the impact of reducing the rescaler multiplicand bit-width on hardware cost by means of a commercial embedded-NPU architecture, we synthesize the systolic dot-product-array including the rescaler from \cite{neutron_ref} in a commercial 16-nm technology. As shown in Figure~\ref{fig:rescaling-unit}, scaling down from 32 to 4 bits yields area reductions of 58.5\%, 47.5\%, and 34.8\% for dot-product-engines deploying 4, 8, and 16 parallel MACs, respectively. These savings solely stem from reduced rescaling multiplicand width. 

%which reduces the cost of the wide rescaling multiplier hardware but also its output width and thereby the cost of the following shift-round-and-saturate stage.

%which lowers the number of partial products and overall logic complexity.

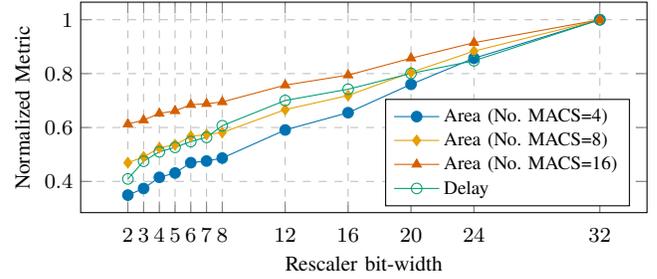
\begin{figure}[t]
    \centering
    \begin{tikzpicture}
\footnotesize
  \begin{axis}[
        grid=major,
        xlabel={Rescaler bit-width},
        ylabel={Normalized Metric},
        width=0.85\linewidth,
        height=2.8cm,
        legend pos=south east,
        legend style={cells={anchor=west}, font=\small},
        xtick={2,3,4,5,6,7,8,12,16,20,24,32},
        ytick={0.2,0.4,0.6,0.8,1.0},
        yticklabel style={
            /pgf/number format/fixed,
            /pgf/number format/fixed zerofill=false,
            /pgf/number format/1000 sep={}
        },
        ymajorgrids=true,
        grid style=dashed,
        legend style={nodes={scale=0.8, transform shape}},
        xticklabel style={yshift=-0.1cm}
    ]

    \pgfplotstableread[col sep=space]{data/rescale_area_saving.data}\datatable

    % Plot MACS = 4
    \addplot[mark=*,blue] 
      table[x=BW, y=MACS4] {\datatable};
    \addlegendentry{Area (No. MACS=4)}

    % Plot MACS = 8
    \addplot[mark=diamond*,orange] 
      table[x=BW, y=MACS8] {\datatable};
    \addlegendentry{Area (No. MACS=8)}

    % Plot MACS = 16
    \addplot[mark=triangle*,red] 
      table[x=BW, y=MACS16] {\datatable};
    \addlegendentry{Area (No. MACS=16)}

    \addplot[mark=o,green] 
      table[x=BW, y=delay] {\datatable};
    \addlegendentry{Delay}

  \end{axis}
\end{tikzpicture}
    \vspace{-2mm}
    \caption{Silicon area of the processing unit from \autoref{fig:rescaling-unit}. %evaluated across varying numbers of MAC units and rescaler bit-widths in a commercial 16-nm technology.
    }
    \vspace{-2mm}
    \label{fig:rescale_area_saving}
\end{figure}

Relative power savings correlate with area reductions. Timing analysis shows nearly 50\% improvement in the critical path across all MAC trees—comprising multiplication, adder tree, accumulation, and rescaling operations—since the rescaler multiplication consistently determines the critical path. Overall, the proposed approach reduces the area-delay product of integer-only AI accelerators by over $4\times$. For 8-bit rescale-multiplicands (to not require fine-tuning) instead of 4, all savings are approximately halved, implying still an area-delay product improvement by over 2$\times$. 

%The resulting hardware not only consumes significantly less energy—as low as 0.2pJ per \verb|INT8| multiply versus 3.1pJ for \verb|INT32| \cite{horowitz2014}—but reduces the propagation delay as well.  
%also benefits from lower latency due to reduced propagation delay and fewer logic stages.
\vspace{-2mm}
\section{Conclusion}
\vspace{-1mm}
This work shows that rescale-aware quantization and incremental training enable highly efficient embedded NPU architectures. Reducing the rescaler bit-width from 32 to 8 bits yields over $2\times$ area-delay product improvement without accuracy loss over standard \textit{W8A8} model quantization. With lightweight fine-tuning, even aggressive 4-bit rescaler quantization achieves over $4\times$ improvement while preserving model quality. Rescaling optimization thus offers a promising path to lower complexity and energy consumption in quantized inference. In future work, we will extend this approach to emerging Gen-AI and Agentic-AI applications at the edge.
% This work demonstrates the effectiveness of rescale-aware quantization and incremental training in enabling ultra-resource-efficient deployment of neural networks. Specifically, reducing the bit-width of the rescaling multiplicand from 32-bit to 8-bit post-training yields an over $2\times$ improvement in the area-delay product at no model quality loss over standard \textit{W8A8} model quantization.
% To fully recover the accuracy loss even for aggressive 4-bit rescaler quantization---delivering over $4\times$ area-delay product improvements--- a light-weight rescale-aware fine-tuning was introduced. 

% Thus, rescaling optimization is a promising approach to reducing the complexity and energy consumption of quantized inference computing pipelines without sacrificing accuracy. In future work, we will hence extend this approach to emerging Gen-AI and Agentic-AI applications at the edge.
\vspace{-2mm}
%—particularly when retraining is feasible.
% \input{sections/future_work}
% \input{sections/acknowledgement}

% \color{blue}
% \begin{itemize}
%     \item Motivation (potential of optimization in the rescaler)
%     \item Approach of fine-tuning (3/4 model paradigm)
%     \item Validation of data output between tflite and own keras model (Sanity Check)
%     \item Presentation of results (potential gain)
% \end{itemize}

%\vfill\null % move refs to own column
\bibliographystyle{bib/IEEEtran}
% \balance
\bibliography{bib/references}

\end{document}